\documentclass[conference]{IEEEtran}
\IEEEoverridecommandlockouts

\usepackage{cite}
\usepackage{amsmath,amssymb,amsfonts}
\usepackage{algorithmic}
\usepackage{graphicx}
\usepackage{textcomp}
\usepackage{booktabs}

\usepackage{tabularx}
\usepackage{multirow}
\usepackage[caption=false,font=normalsize,labelfont=sf,textfont=sf]{subfig}
\usepackage[table,xcdraw]{xcolor}

\usepackage{pifont}
\newcommand{\cmark}{\text{\ding{51}}}
\newcommand{\xmark}{\text{\ding{55}}}

\def\BibTeX{{\rm B\kern-.05em{\sc i\kern-.025em b}\kern-.08em
    T\kern-.1667em\lower.7ex\hbox{E}\kern-.125emX}}
\begin{document}

\title{Improving Noise Robust \\ Audio-Visual Speech Recognition \\ via Router-Gated Cross-Modal Feature Fusion}

\author{
\IEEEauthorblockN{ DongHoon Lim\textsuperscript{*}, YoungChae Kim\textsuperscript{*}, Dong-Hyun Kim, Da-Hee Yang, Joon-Hyuk Chang\textsuperscript{\dag}}
\IEEEauthorblockA{ \textsuperscript{*}\textit{Dept. Artificial Intelligence}, \textit{Hanyang University}, Seoul, Republic of Korea \\
}
\IEEEauthorblockA{ \textit{Dept. Electronic and Electrical Engineering}, \textit{Hanyang University}, Seoul, Republic of Korea \\
\{hswdh, yc0604, dlstjd1477, douxi15, jchang\}@hanyang.ac.kr}
\thanks{\textsuperscript{*}Co-first authors, equal contribution
\textsuperscript{\dag}Corresponding author.}
}

\maketitle

\section*{}

\begin{abstract}
Robust audio-visual speech recognition (AVSR) in noisy environments remains challenging, as existing systems struggle to estimate audio reliability and dynamically adjust modality reliance. We propose router-gated cross-modal feature fusion, a novel AVSR framework that adaptively reweights audio and visual features based on token-level acoustic corruption scores. Using an audio-visual feature fusion-based router, our method down-weights unreliable audio tokens and reinforces visual cues through gated cross-attention in each decoder layer. This enables the model to pivot toward the visual modality when audio quality deteriorates. Experiments on LRS3 demonstrate that our approach achieves an 16.51–42.67\% relative reduction in word error rate compared to AV-HuBERT. Ablation studies confirm that both the router and gating mechanism contribute to improved robustness under real-world acoustic noise.
\end{abstract}


\begin{IEEEkeywords}
Audio-Visual Speech Recognition, Router-Gated Cross Attention, Noise-Robust ASR, Cross-Modal Fusion
\end{IEEEkeywords}

\begin{figure*}[!tbp] 
    \centering
    \includegraphics[width=\textwidth]{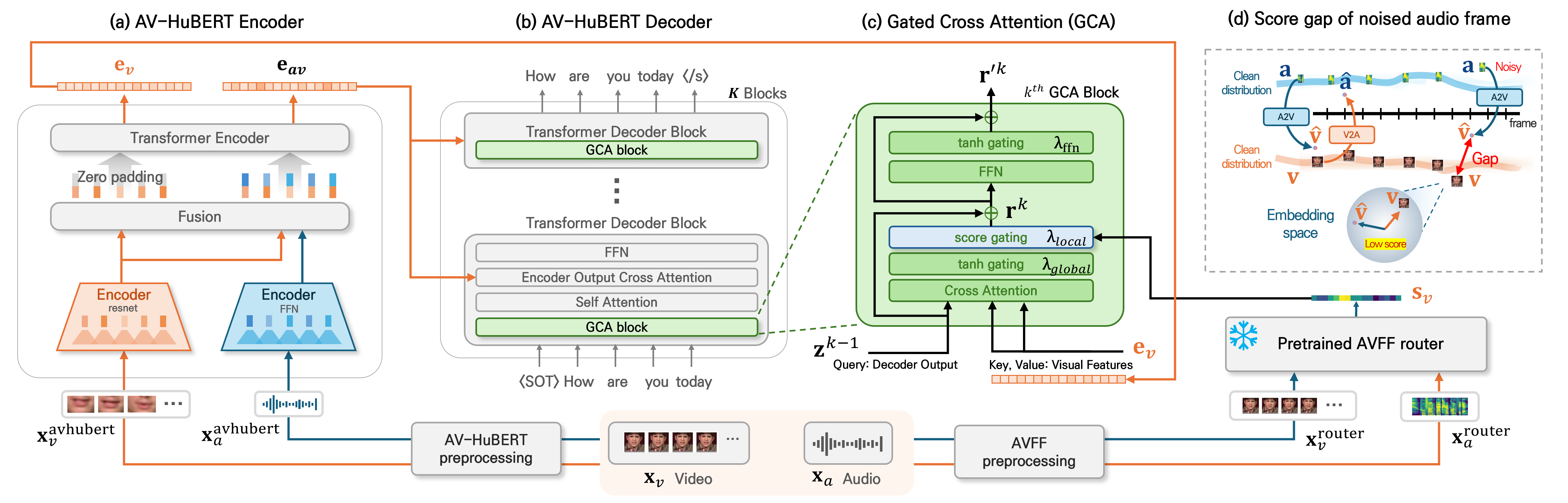}
    \caption{The diagram illustrates the overall architecture, extending the AV-HuBERT encoder-decoder structure by integrating a pre-trained AVFF-based router into each decoder block. As shown in (d), the gap between actual and predicted embeddings (e.g., $\mathbf{v}$ vs. $\hat{\mathbf{v}}$ from noisy $\mathbf{a}$) reflects the reliability score, with larger gaps indicating lower reliability.}
    \label{figure_1}
    \vspace{-1em}
\end{figure*}

\section{Introduction}
Audio-visual speech recognition (AVSR) aims to improve speech recognition by integrating acoustic and visual information, typically from the speaker’s lips. This multimodal approach is especially valuable in noisy environments, where audios are degraded but visual cues remain intact. However, existing AVSR systems still rely heavily on the audio stream\cite{lin2025uncovering}, leading to limited performance gains in noisy conditions. For example, models such as AV-HuBERT \cite{avhubert} and Whisper-Flamingo\cite{whisper-flamingo} remain audio-centric, showing only marginal gains over audio-only systems, even under significant noise.

This imbalance stems from how these systems handle cross-modal fusion. While many architectures combine audio and visual cues---either through encoder-level fusion or decoder-level gated cross-attention (GCA)---they often fail to adapt modality weighting to input reliability. For instance, Whisper-Flamingo applies a scalar gate per decoder layer to modulate visual input, but this gating mechanism is independent of audio quality. Moreover, AVSR systems are typically biased toward the acoustic modality, as they are often trained on only audio data. As noted in~\cite{watchorlisten}, recognizers trained on clean conditions tend to rely heavily on audio, since it provides sufficient information for accurate speech recognition, ultimately underutilizing visual cues when noise corrupts the audio input. Prior efforts have attempted to address this issue by incorporating noise awareness into AVSR. These include uncertainty-aware attention\cite{nevertolate}, modality dropout\cite{avhubert}, and  signal-to-noise ratio (SNR) prediction\cite{snr_pred}. However, these methods operate at a broader temporal granularity (e.g., frame or clip level), where each decision aggregates information over multiple embedding tokens, potentially losing fine-grained alignment across modalities.


In this work, we propose \textit{Router-Gated Cross-Modal Feature Fusion}, a noise-adaptive decoding framework that dynamically modulates visual reliance based on localized acoustic quality. Our approach leverages the audio-visual feature fusion (AVFF)\cite{avff} module as a router that computes token-level similarity between audio and visual embeddings without explicit noise labels. A low similarity score indicates corrupted audio, prompting the decoder to increase reliance on visual features via residual GCA blocks. This allows the model to adaptively shift attention to the visual modality when needed, while maintaining acoustic dominance in clean regions.

Unlike previous works, our method operates at the fine-grained token level and supports layer-wise adaptation. It is compatible with existing Transformer-based decoders---including AV-HuBERT and Whisper-Flamingo---and can be integrated as a plug-in module without retraining the backbone on specific noise types. This design enables robust, context-sensitive fusion across a wide range of acoustic conditions. Our contributions are threefold:
\begin{enumerate}
\item We introduce a router-guided attention mechanism that dynamically reallocates modality focus at the token and layer level, inspired by the human ability to shift attention based on input quality.
\item We present the first AVSR decoder architecture that integrates token-level audio reliability into every GCA block, enabling context-sensitive visual enhancement.
\item We demonstrate consistent word error rate reductions of 16.51–42.67\% on LRS3 under various noise conditions, surpassing AV-HuBERT and confirming the benefit of dynamic and noise-aware fusion.
\end{enumerate}

\section{\textbf{Related Works}}

\subsection{Audio-Visual Fusion in Speech Recognition}
Audio-visual speech recognition (AVSR) has gained increasing attention due to its potential to improve robustness under adverse acoustic conditions by incorporating visual cues such as lip movements \cite{avsris, RNNT}. Recent works, including AV-HuBERT\cite{avhubert} and AV-data2vec \cite{avdata2vec}, have leveraged self-supervised learning (SSL) to jointly learn representations from audio and visual modalities. These approaches have demonstrated impressive gains in recognition accuracy, particularly under clean or moderately noisy conditions. However, despite the use of bimodal inputs, most of these models remain inherently audio-centric in both architecture and training objectives\cite{robustssl}. As a result, they often fail to effectively leverage visual information precisely when most critical---under severe acoustic degradation \cite{watchorlisten}.


\subsection{Noise Robustness in AVSR}
To mitigate performance degradation under acoustic noise, various strategies have been proposed. Data-centric approaches such as MUSAN\cite{noise} and ESC-50\cite{ESC} introduce noise mixtures during training\cite{avhubert}, while augmentation techniques like SpecAugment perturb spectrograms to improve generalization\cite{specaugment, dcimavsr}. Complementarily, speech enhancement front-ends like DCCRN\cite{DCCRN} and Demucs\cite{demucs} have been deployed to denoise audio before passing it to recognition models. While these techniques can be effective when the noise characteristics are well matched between training and testing, they often exhibit limited generalization to unseen noise types or domain shifts, and frequently require extensive retraining with labeled datasets to maintain performance.


\subsection{Architectural Approaches to Visual Integration}
Beyond preprocessing, architectural modifications have been introduced to better incorporate visual information into AVSR systems. Encoder-level strategies include multi-stream Conformer\cite{dcimavsr} architectures and hybrid CNN–Transformer models that jointly encode visual and auditory features\cite{avsrconformers}. On the decoder side, models such as Whisper-Flamingo \cite{whisper-flamingo} inject visual embeddings into attention mechanisms to influence token prediction. However, a common limitation across these approaches is that fusion mechanisms are typically static---relying on global scalar gates or uniform cross-modal attention. Such fusion lacks the flexibility to adapt modality weights based on local acoustic conditions, leading to suboptimal reliance on visual cues when the auditory is unreliable\cite{weighting}.


\subsection{Dynamic and Reliability-Aware Fusion}
Recognizing the limitations of static fusion, recent work has explored dynamic, reliability-aware mechanisms. These include uncertainty-aware attention\cite{nevertolate}, SNR-conditioned weighting\cite{snr_pred}, and modality dropout\cite{avhubert} during training to encourage robustness against missing or degraded inputs. While these techniques improve adaptability to some extent, most operate at the frame or clip level and use shallow reliability predictors, limiting their ability to perform token-level decisions that align with fine-grained acoustic context. Moreover, few approaches are designed to modulate modality weighting across network depth, which can be an important consideration in hierarchical Transformer models\cite{unibev}.

\section{\textbf{Method}}
We propose a token-level modality fusion mechanism that dynamically adjusts the contributions of audio and visual modalities based on reliability scores predicted by a pretrained AVFF router. This design is inspired by findings in cognitive science and neuroscience, which show that humans adaptively shift modality reliance based on sensory reliability~\cite{mcgurk, flexible}. As illustrated in Fig. 1, the AVFF router computes frame-wise audio reliability from raw inputs, which are then injected as gating factors into GCA layers embedded in the Transformer decoder of an AV-HuBERT-based AVSR model. The overall system consists of (1) an AV-HuBERT encoder–decoder backbone, (2) an AVFF router that estimates audio reliability, and (3) GCA blocks with gated routing based on local and global control.

\begin{figure*}[!tbp] 
    \centering
    \includegraphics[width=\textwidth]{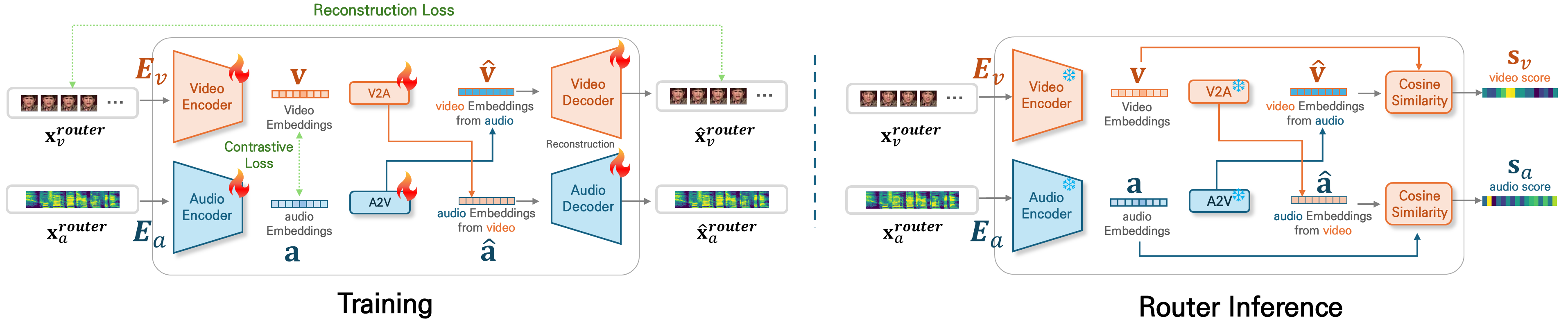}
    \caption{AVFF pre-training and router inference process, which enables the AVFF-based router to detect modality imbalance. During the inference stage, with the decoder removed, modality reliability is determined solely by cosine similarity.}
    \label{figure_2}
    \vspace{-1em}
\end{figure*}

\subsection{Overview}
Our model is built upon AV-HuBERT, a masked prediction-based audio-visual speech representation model. Given audio-visual input pairs $(\mathbf{x}_a, \mathbf{x}_v)$, the AV-HuBERT encoder transforms them into fused hidden representations (Fig. 1(a)). These are then passed to a Transformer decoder trained from scratch (Fig. 1(b)), with the encoder initialized from pretrained AV-HuBERT. We enhance this decoder with GCA blocks inserted at each layer to enable dynamic fusion of audio and visual information (Fig. 1(c)). Each GCA block integrates visual context into the audio decoding path, with the degree of integration governed by token-level reliability scores (Fig. 1(d)) from the separately pretrained AVFF router. This enables the decoder to increase visual reliance in noisy segments while reducing reliance on the visual modality when the audio signal is clean.

\subsection{AV-HuBERT}
Following AV-HuBERT’s canonical preprocessing, we obtain a lip-synchronized video tensor $\mathbf{x}^{\text{avhubert}}_{v}$ and a 16 kHz waveform segment $\mathbf{x}^{\text{avhubert}}_{a}$. These two streams are fed into the model’s uni-modal front-end ResNet towers for vision and a 1-D convolutional stack for audio—which generate the initial visual and acoustic feature sequences. The resulting features are then passed to the Transformer encoder, which produces a sequence of fused representations. Depending on the input modality configuration, the encoder yields either of the following embeddings:
\begin{itemize}
    \item Audio–visual embedding $\mathbf{e}_{av}$: obtained when both $\mathbf{x}^{\text{avhubert}}_{v}$ and $\mathbf{x}^{\text{avhubert}}_{a}$ are provided.
    \item Visual-only embedding $\mathbf{e}_{v}$: obtained when only $x^{\text{avhubert}}_{v}$ is input and the audio branch is zero-masked.
\end{itemize}
The fused audio-visual embedding $\mathbf{e}_{av}$ is forwarded to the subsequent Transformer decoder as its encoder memory, whereas the visual-only embedding $\mathbf{e}_{v}$ is cached as the key/value source for the GCA blocks.

\subsection{AVFF-Based Router}
Fig.~\ref{figure_2} illustrates the overall architecture of our AVFF-based token router, which estimates token-wise audio reliability by leveraging cross-modal reconstruction signals. 

\noindent\textbf{Model Overview.}  
The AVFF model comprises two modality-specific masked autoencoders (MAEs) for audio and video, along with two cross-modal translators—A2V (audio-to-video) and V2A (video-to-audio). These translators aim to reconstruct the latent representation of one modality using information from the other. Both A2V and V2A are composed of a linear projection followed by the Transformer layer that maps one modality's representation into the space of the other.

\noindent\textbf{Training Phase }(Left pane of Fig.~\ref{figure_2}).  
Given raw waveform and video frames, we apply AVFF preprocessing to obtain synchronized inputs: $\mathbf{x}^{\mathrm{router}}_{a}$ for audio and $\mathbf{x}^{\mathrm{router}}_{v}$ for video. These inputs are patchified and fed into the audio and video MAE encoders, $E_a$ and $E_v$, yielding latent representations:
\vspace{-0.2em}
\begin{equation}
   \mathbf{a} = E_a(\mathbf{x}^{\mathrm{router}}_{a}), \qquad \mathbf{v} = E_v(\mathbf{x}^{\mathrm{router}}_{v}).
   \vspace{-0.2em}
\end{equation}
To facilitate cross-modal learning, each translator produces a prediction of one modality’s latent from the other:
\vspace{-0.2em}
\begin{equation}
   \hat{\mathbf{v}} = \text{A2V}(\mathbf{a}), \qquad \hat{\mathbf{a}} = \text{V2A}(\mathbf{v}),
   \vspace{-0.2em}
\end{equation}

where $\hat{\mathbf{v}}$ approximates the visual latent features reconstructed solely from audio, and $\hat{\mathbf{a}}$ approximates the audio latent features reconstructed solely from video.


\noindent\textbf{Inference Phase }(Right pane of Fig.~\ref{figure_2}). 
During inference, only the encoders and translators are retained, and the MAE decoders are discarded. To assess the reliability of each modality, we compute the cosine similarity between the original and cross-modal predicted embeddings:
\vspace{-0.2em}
\begin{equation}
    \mathbf{s}_{v} = \cos(\mathbf{v}, \hat{\mathbf{v}}), \qquad \mathbf{s}_{a} = \cos(\mathbf{a}, \hat{\mathbf{a}}),
    \vspace{-0.2em}
\end{equation}
\noindent where $\cos(\cdot,\cdot)$ denotes the cosine similarity function. Since $\hat{\mathbf{v}}$ is derived from audio only, $\mathbf{s}_{v}$ serves as a proxy for audio reliability: lower values indicate acoustic corruption or audio-visual misalignment. The visual reliability score $\mathbf{s}_{v}$ is transformed into a visual attention gain by computing its linearly shifted value, $(1-\mathbf{s}_{v})$, thereby increasing visual emphasis when reliability is low. Subsequently, the shifted value $(1-\mathbf{s}_{v})$ is linearly interpolated\cite{align_linearinterpolation} to match the AV-HuBERT token sequence of length $N$. As a result, this interpolated gain is used for token-wise modulation of visual features.


\vspace{-0.2em}
\begin{equation}
    \boldsymbol{\lambda}_{{local}} = \tanh\bigl(\text{Interpolate}(1 - \mathbf{s}_{v})\bigr),
    \vspace{-0.2em}
\end{equation}
where $\boldsymbol{\lambda}_{\text{local}} \in \mathbb{R}^N$ is a token-wise gating factor applied to the visual embedding stream in each GCA block. This factor enhances the influence of visual cues when the audio is unreliable, and attenuates them when the audio is trustworthy.

\subsection{Gated Cross-Attention Scoring}

We enhance each decoder block in AV-HuBERT by inserting a GCA block, as shown in Fig.~\ref{figure_1}(b), inspired by the Flamingo \cite{flamingo}. This module is placed before the self-attention, encoder cross-attention, and feed-forward (FFN) submodules. The GCA block selectively injects more visual information into the decoder hidden state, especially when the audio signal is deemed unreliable. Let $K$ denote the number of decoder layers, and let $\mathbf{z}^k \in \mathbb{R}^{N \times F}$ denote the hidden state at layer $k$, where $N$ is the sequence length and $F$ is the feature dim.

\noindent\textbf{Cross-modal Attention.}  
At each layer $k$, we first compute the cross-attention output between the layer normalized previous decoder hidden
state $\text{LN}(\mathbf{z}^{k-1})$ and the $\mathbf{e}_v \in \mathbb{R}^{N \times F}$:
\vspace{-0.5em}

\begin{equation}
   \mathbf{A}^k = \text{Attn}(\text{LN}(\mathbf{z}^{k-1}), \mathbf{e}_v),
   \vspace{-0.2em}
\end{equation}

\noindent where ``LN" denotes layer-normalization\cite{layernorm} and ``Attn" denotes multi-head cross-attention. This captures visual context relevant to the current decoder state.

\noindent\textbf{Gating Mechanisms.}  
To modulate the contribution of the visual signal, we introduce two gating factors:

\begin{itemize}
    \item \textit{Global gate} ($\boldsymbol{\lambda}_{\text{global}}$): A learnable scalar $\alpha$ per GCA block, transformed via $\tanh$, $\boldsymbol\lambda_{global} = \tanh(\alpha)$, that controls the overall strength of the visual injection.
    
    \item \textit{Local gate} ($\boldsymbol{\lambda}_{\text{local}}$): A token-wise reliability vector computed from the AVFF-based router, indicating the reliability of each audio token. 
\end{itemize}

The visual cross-attention output $\mathbf{A}^k$ is modulated by both gates and added to the decoder state:
\vspace{-0.5em}

\begin{equation}
    \mathbf{r}^k = \mathbf{z}^{k-1} + \boldsymbol{\lambda}_{\text{global}} \cdot \boldsymbol{\lambda}_{\text{local}} \cdot \mathbf{A}^k,
    \vspace{-0.2em}
\end{equation}
where \(\mathbf{r}^k\) denotes the intermediate representation in the \(k\)-th GCA block, obtained by adding gated visual cross-attention output to the previous decoder hidden state \(\mathbf{z}^{k-1}\). This intermediate representation is then passed through a feed-forward network (FFN) to produce the updated GCA block output \(\mathbf{r}^{\prime k}\).

\noindent\textbf{GCA Block Output.}  
The FFN module follows, with an additional scalar gate $\boldsymbol{\lambda}_{\text{ffn}} = \tanh(\alpha_{\text{ffn}})$ applied to modulate its contribution, consistent with the approach in Flamingo. Here, $\alpha_{\text{ffn}}$ is a learnable scalar that controls the influence of the FFN output at each GCA block.

\vspace{-0.5em}

\begin{equation}
   \mathbf{r}^{\prime k} = \mathbf{r}^{k} + \boldsymbol{\lambda}_{\text{ffn}} \cdot \text{FFN}(\text{LN}(\mathbf{r}^k)).
   \vspace{-0.5em}
\end{equation}

This gated formulation allows each decoder layer to adaptively rely on visual cues based on audio reliability at both the global and local levels, ensuring robustness while preserving stability at initialization. In this study, we restrict noise corruption to the audio stream and assume clean, synchronized video.

\begin{table*}[!tbp]
\centering
\caption{WER (\%) and RERR (\%) result on LRS3 (in-domain) and LRS2 (out-of-domain) test sets under clean and noisy conditions. Following pretraining on clean VoxCeleb2+LRS3, models were fine-tuned on LRS3 433h (clean condition) and LRS3 433h augmented with MUSAN random noise (0 dB SNR) (noise condition).}
\resizebox{\textwidth}{!}{%
\begin{tabular}{ccccccccccccccc}
\toprule[1.5pt]
                                  &                                                                                                                                                         & \multicolumn{1}{c}{}                                                                       &                                  & \multicolumn{3}{c}{\textbf{Babble, SNR (dB)}}           & \multicolumn{3}{c}{\textbf{Music, SNR (dB)}} & \multicolumn{3}{c}{\textbf{Natural, SNR (dB)}} & \multicolumn{2}{c}{\textbf{avg}}                               \\   \cmidrule(r){5-13} \cmidrule(r){14-15} 
\rule{0pt}{2ex} \multirow{-3}{*}{\textbf{Method}}  & \multirow{-3}{*}{\textbf{\begin{tabular}[c]{@{}c@{}}FT type\end{tabular}}} & \multicolumn{1}{c}{\multirow{-3}{*}{\textbf{\begin{tabular}[c]{@{}c@{}}test \\
domain\end{tabular}}}} & \multirow{-3}{*}{\textbf{clean}} &  \textbf{10}                   & \textbf{5} & \textbf{0} & \textbf{10}    & \textbf{5}   & \textbf{0}   & \textbf{10}    & \textbf{5}    & \textbf{0}    & \textbf{WER (\%)} & \textbf{RERR (\%)}\\  \toprule[1.5pt]

AV-HuBERT                                                                                                &                      &                                                    & 3.88     & 15.07 & 20.12      & 35.92      & 8.12           & 10.05        & 12.84        & 6.94           & 8.48          & 12.92         & 13.43    &-                            \\

AV-HuBERT (Ours)                                                                       &    \multirow{-2}{*}{\begin{tabular}[c]{@{}c@{}}\textbf{clean}\\ (LRS3)\end{tabular}}                 & \multirow{-2}{*}{\begin{tabular}[c]{@{}c@{}}LRS3\end{tabular}}                                                    & \textbf{3.09}                         & \textbf{6.80}                           & \textbf{9.19}       & \textbf{21.40}       & \textbf{3.88}           & \textbf{5.31}         & \textbf{8.96}         & \textbf{4.75}           & \textbf{5.47}          & \textbf{8.16}          &   \textbf{7.70}                   &\textbf{42.67}          \\ \hline

\rule{0pt}{2.5ex} AV-HuBERT                                                                                                 &          &                                                                & 5.47     & 7.01  & 8.56       & 14.11      & 7.91           & 8.84         & 10.35        & 6.85           & 7.61          & 9.24          & 8.60                  &-            \\
AV-HuBERT (Ours)                                                                       &    \multirow{-2}{*}{\begin{tabular}[c]{@{}c@{}}\textbf{noise}\\ (LRS3+MUSAN)\end{tabular}}                                  &           \multirow{-2}{*}{\begin{tabular}[c]{@{}c@{}}LRS3\end{tabular}}                          & \textbf{5.33} & \textbf{6.42} & \textbf{7.45} & \textbf{12.68} & \textbf{5.95} & \textbf{6.66} & \textbf{8.32}  & \textbf{5.63} & \textbf{6.18}  & \textbf{7.21}               &  \textbf{7.18}            & \textbf{16.51}             \\ \hline

\rule{0pt}{2.5ex}AV-HuBERT                                                                                                  &             &                                                             &  16.16    & 21.42 &   30.07    &   43.97    &  16.78          & 19.56        & 21.73        &  18.22         &  21.05        &  26.46       & 23.54                     &-           \\
AV-HuBERT (Ours)                                                                     &    \multirow{-2}{*}{\begin{tabular}[c]{@{}c@{}}\textbf{clean}\\ (LRS3)\end{tabular}}         &    \multirow{-2}{*}{\begin{tabular}[c]{@{}c@{}}LRS2\end{tabular}}                                                &  \textbf{14.11}    & \textbf{17.56} &  \textbf{22.50}     & \textbf{35.90}      &  \textbf{15.05}          &  \textbf{16.60}       &  \textbf{20.30}       &  \textbf{15.75}         &  \textbf{16.80}        &  \textbf{20.30}       & \textbf{19.49}     & \textbf{17.20}                     \\ \hline

\rule{0pt}{2.5ex} AV-HuBERT                                                                                                  &          &                                                               & 16.58     & 19.05 & 21.22      &  \textbf{24.09}     &    17.09        &  17.91       &   19.26      & \textbf{17.09}          &  18.50        &  19.25        & 19.00                      &-         \\
AV-HuBERT (Ours)                                                                       &    \multirow{-2}{*}{\begin{tabular}[c]{@{}c@{}}\textbf{noise}\\ (LRS3+MUSAN)\end{tabular}}                         &   \multirow{-2}{*}{\begin{tabular}[c]{@{}c@{}}LRS2\end{tabular}}                                           & \textbf{16.20} & \textbf{16.92} & \textbf{19.60} & 25.00 &  \textbf{16.45} & \textbf{16.68} & \textbf{18.55} & 17.50 & \textbf{17.96} & \textbf{19.17}              &    \textbf{18.40}            &\textbf{3.15}               \\
 \toprule[1.5pt]
\end{tabular}
}
\label{tab:comparison}
\end{table*}

\begin{table*}[!tbp]
\centering
\caption{WER (\%) and RERR (\%) comparisons on the LRS3 (in-domain) and LRS2 (out-of-domain) benchmarks using Whisper-Flamingo, with and without the proposed AVFF router. Following noise pretraining on LRS3, models were fine-tuned on LRS3 433h (clean condition) augmented with MUSAN noise (0 dB SNR) (noise condition).}
\resizebox{\textwidth}{!}{%
\begin{tabular}{ccccccccccccccc}
\toprule[1.5pt]
                                  &                                                                                                                                                         & \multicolumn{1}{c}{}                                                                       &                                  & \multicolumn{3}{c}{\textbf{Babble, SNR (dB)}}           & \multicolumn{3}{c}{\textbf{Music, SNR (dB)}} & \multicolumn{3}{c}{\textbf{Natural, SNR (dB)}} & \multicolumn{2}{c}{\textbf{avg}}                               \\   \cmidrule(l){5-13} \cmidrule(l){14-15}
\rule{0pt}{2ex} \multirow{-3}{*}{\textbf{Method}}  & \multirow{-3}{*}{\textbf{\begin{tabular}[c]{@{}c@{}}FT type\end{tabular}}} & \multicolumn{1}{c}{\multirow{-3}{*}{\textbf{\begin{tabular}[c]{@{}c@{}}test \\
domain\end{tabular}}}} & \multirow{-3}{*}{\textbf{clean}} &  \textbf{10}                   & \textbf{5} & \textbf{0} & \textbf{10}    & \textbf{5}   & \textbf{0}   & \textbf{10}    & \textbf{5}    & \textbf{0}    & \textbf{WER (\%)} & \textbf{RERR (\%)}\\  \toprule[1.5pt]

Whisper-flamingo                                                                                          &                                                                    &      & 2.62                         & 3.37                          & 5.15       & 13.01         & 2.84           & 3.12         & 3.91         & 3.09           & 3.36          & 4.44         & 4.49       &-                  \\
Whisper-flamingo (Ours)                       &    \multirow{-2}{*}{\begin{tabular}[c]{@{}c@{}}\textbf{noise}\\ (LRS3+MUSAN)\end{tabular}}                 & \multirow{-2}{*}{\begin{tabular}[c]{@{}c@{}}LRS3\end{tabular}}        & \textbf{2.04}          & \textbf{2.70}                           & \textbf{4.40}        & \textbf{12.30}       & \textbf{2.43}           & \textbf{2.81}         & \textbf{3.74}         & \textbf{2.22}           & \textbf{2.65}          & \textbf{3.19}          & \textbf{3.85}                  &\textbf{14.25}        
\\ \hline
\rule{0pt}{2.5ex} Whisper-flamingo                                                                                          &                                               &                          & 12.77     & 14.64 & 17.69      &  29.42     & 13.63           & 13.85        & 15.64        & 13.76          & 14.60         & 17.02        & 16.30           &-           \\
Whisper-flamingo (Ours)                       &    \multirow{-2}{*}{\begin{tabular}[c]{@{}c@{}}\textbf{noise}\\ (LRS3+MUSAN)\end{tabular}}                 & \multirow{-2}{*}{\begin{tabular}[c]{@{}c@{}}LRS2\end{tabular}}                                                                   & \textbf{12.02}     & \textbf{13.67}  &  \textbf{16.39}     & \textbf{27.90}      & \textbf{12.67}           & \textbf{13.25}        &   \textbf{14.70}      &  \textbf{12.88}         &  \textbf{13.28}        & \textbf{15.60}        &  \textbf{15.24}  &\textbf{6.50}                     
\\ \toprule[1.5pt]

\end{tabular}%
}
\vspace{-1.5em}
\label{tab:ood}
\end{table*}


\subsection{Training Objectives and Optimization}
\noindent\textbf{Pretraining.} Following \cite{avff}, our AVFF router (Fig. 2) is pretrained using a combination of contrastive, reconstruction, and adversarial losses to promote cross-modal alignment and robust distribution modeling. Concretely, a contrastive loss $\mathcal{L}_c$ is applied to the video $\mathbf{v}$ and audio $\mathbf{a}$ embeddings to encourage modality-invariant representation learning. A reconstruction loss $\mathcal{L}_r$ is employed to minimize the mean square error (MSE) between the MAE decoder outputs $(\hat{\mathbf{x}}^{\text{router}}_a, \hat{\mathbf{x}}^{\text{router}}_v)$ and the original inputs $(\mathbf{x}^{\text{router}}_a, \mathbf{x}^{\text{router}}_v)$, preserving autoencoding capabilities. Additionally, a Wasserstein generative adversarial networks (WGAN)\cite{wgan} loss $\mathcal{L}_{adv}$ is used to enhance the stability of high-dimensional representation learning through adversarial training. The total pretraining objective is formulated as a weighted sum:
\vspace{-0.5em}
\begin{equation}
\mathcal{L}_{\text{total}} = \lambda_c \mathcal{L}_c + \lambda_r \mathcal{L}_r + \lambda_{adv} \mathcal{L}_{adv},
\label{eq:loss}
\vspace{-0.5em}
\end{equation}
\noindent where $\lambda_c$, $\lambda_r$, and $\lambda_{adv}$ are the weighting coefficient.

\noindent\textbf{Fine-tuning.} During the fine-tuning phase, the pretrained AVFF router remains frozen, while the AV-HuBERT model---including the encoder, Transformer decoder, GCA modules, and all gating parameters---is fine-tuned end-to-end using a sequence-to-sequence cross-entropy (CE) loss. The CE loss is computed between the decoder outputs and the ground-truth transcriptions, and the model is optimized using the Adam optimizer~\cite{Adam}. To enhance training stability and mitigate overfitting, the encoder is kept frozen for the first $\tau$\ steps.

\section{\textbf{Experiments}}

\subsection{Experimental Setup}

\noindent\textbf{Datasets.} We conducted our experiments using two publicly available audio-visual speech datasets: LRS3 and LRS2. The LRS3 dataset comprises a large-scale collection of English speech clips extracted from TED and TEDx talks\cite{lrs3}, while LRS2 consists of approximately 224 hours of broadcast speech from BBC programs\cite{lrs2}. To evaluate model robustness under noisy conditions, we applied audio corruptions using the MUSAN corpus, incorporating three types of background noise---babble, music, and natural. Corrupted audio samples were generated at signal-to-noise ratios (SNRs) of 10 dB, 5 dB, and 0 dB.

\noindent\textbf{Preprocessing.} For AV-HuBERT, the audio stream was converted into a 26-dimensional log filterbank energy representation with a 10 ms stride. The video stream was cropped to a $96 \times 96$ region-of-interest (ROI) centered on the mouth. During training, a random $88 \times 88$ crop was applied to the ROI, while inference used a center crop of the same size. For the AVFF Router, the audio stream was transformed into a Mel-spectrogram with 128 bins, using a 16 ms Hamming window and a 4 ms hop size. Video inputs were segmented into fixed-duration clips of 3.2 seconds by uniformly sampling frames from the original sequence. Each clip was then processed to yield 768 audio frames and 16 visual frames, subsequently divided into non-overlapping patches: visual frames into $2 \times 16 \times 16$, and audio spectrograms into $16 \times 16$.


\noindent\textbf{Implementation Details.}
For the AV-HuBERT, we employed a pretrained model publicly available from the \texttt{fairseq} framework. We initialized the encoder using the checkpoint of AV-HuBERT-BASE pretrained on VoxCeleb2 \cite{voxceleb} and LRS3 datasets. The AV-HuBERT decoder was initialized with six Transformer decoder layers, each inherently incorporating self-attention and cross-attention mechanisms designed to attend to the encoder output, and the GCA block was subsequently integrated into each of the decoder layers. All experiments were conducted using 4 NVIDIA RTX 3090 GPUs. 

\noindent\textbf{Pretraining.}
We pretrained the AVFF module on the LRS3 dataset, following the approach proposed in the original study~\cite{avff}. The overall architecture and training pipeline were implemented in accordance with the specifications described in the AVFF paper. For the video encoder and decoder, we adopted MARLIN checkpoints~\cite{MARLIN} pretrained on YouTubeFace, each configured with an embedding dimension of 768. For the audio encoder and decoder, we used AudioMAE checkpoints~\cite{AudioMAE} pretrained on AudioSet-2M, also with an encoder embedding size of 768, while decoder embedding size of 512. These pretrained components served as the backbone for the AVFF router during its self-supervised training phase. The model was trained for 50 epochs with a batch size of 16 using the AdamW optimizer~\cite{adamW}. The loss weights in Eq.~\ref{eq:loss} were set to $\lambda_c = 0.01$, $\lambda_r = 1$, and $\lambda_{adv} = 0.1$.

\begin{table*}[!tbp]
\centering
\caption{Ablation Study Results: Word Error Rate (WER, \%) on the LRS3 benchmark under audio-corrupted conditions with MUSAN noise. All models were fine-tuned on the LRS3 30h dataset. For the clean setting, models were fine-tuned and evaluated on clean data. For the noisy setting, models were fine-tuned with MUSAN random noise at 0 dB SNR and evaluated on test sets corrupted with each corresponding MUSAN noise category (Babble, Music, Natural) at varying SNR levels. The first row indicates the noise type, and the second row shows each corresponding SNR levels(in dB).}
\resizebox{\textwidth}{!}{%
\begin{tabular}{lcccccccccccccc}
\toprule[1.5pt]
\multirow{2}{*}{\textbf{Method}} & \multicolumn{3}{c}{\textbf{Modules}}               & \multirow{2}{*}{\textbf{clean}} & \multicolumn{3}{c}{\textbf{Babble, SNR (dB)}} & \multicolumn{3}{c}{\textbf{Music, SNR (dB)}} & \multicolumn{3}{c}{\textbf{Natural, SNR (dB)}} & \multirow{2}{*}{\textbf{avg}} \\ \cline{2-4} \cline{6-14}
                                 & \textbf{$\mathbf{s}_v$} & \textbf{Cos. Sim.} & \textbf{GCA} &                                 & \rule{0pt}{2.5ex} \textbf{10}   & \textbf{5}   & \textbf{0}     & \textbf{10}    & \textbf{5}   & \textbf{0}   & \textbf{10}    & \textbf{5}    & \textbf{0}    &                               \\ \toprule[1.5pt] 
Baseline                &            &         &      & 6.02                   & 6.80          & 8.00            & 13.48          & 6.10           & 7.37         & 8.10          & 5.94          & 7.30          & 8.56         &   7.77                   \\ \hline
\rule{0pt}{2.5ex} \textbf{Ours}           & \cmark          & \cmark       & \cmark    & \textbf{4.28}          & \textbf{5.39} & \textbf{6.70} & \textbf{11.40} & 5.86             & \textbf{6.10} & 7.60          & \textbf{5.15} & 6.70          & \textbf{7.50} &    \textbf{6.67}                  \\
\multicolumn{1}{r}{$\mathbf{s}_a$}         & \xmark          & \cmark       & \cmark    & 4.99                   & 6.26         & 7.77         & 13.32          & \textbf{5.47} & 6.55         & 7.77         & 5.31          & \textbf{6.20} & 8.10          &   7.17                   \\
\multicolumn{1}{r}{L2} & \xmark          & \xmark       & \cmark    & 4.91                   & 6.50          & 7.77         & 13.30           & 5.80           & 6.60          & \textbf{7.50} & 6.10           & \textbf{6.20}          & 7.70          &   7.25                   \\
\multicolumn{1}{r}{self-attn}                    & \xmark          & \xmark       & \xmark    & 5.07                   & 6.10          & 7.50          & 13.50           & 6.20           & 6.81         & 7.80          & 5.55          & 6.30          & 7.70          &  7.24        \\ \bottomrule[1.5pt]          
\end{tabular}%
}
\vspace{-1em}
\label{tab:abls}
\end{table*}

\noindent\textbf{Fine-tuning} Fine-tuning was conducted under two distinct conditions: clean and noise. For the noise condition, we constructed a fixed dataset by augmenting audio with randomly sampled MUSAN noise (babble, music, natural) at an SNR of 0 dB. The model was fine-tuned for 60K steps on the LRS3 30h subset and 120K steps on the full LRS3 433h dataset. Optimization was performed using the Adam optimizer with a learning rate of 0.001 and a warm-up schedule of 10K steps. The encoder freezing steps $\tau$ were set to the 50\% of the total fine-tuning duration. Additionally, we evaluated the generality of our approach by integrating the proposed score-gating module into the Whisper-Flamingo architecture, inserting it after the tanh gating module of the Whisper-en-x-small checkpoint. For this evaluation, the modified Whisper-en-x-small checkpoint was fine-tuned on the LRS3 433h dataset for 20K steps, following the training setup detailed in the original Whisper-Flamingo paper.

\subsection{Effectiveness of the Proposed Router Across Diverse Conditions}
Table~\ref{tab:comparison} presents the results on the in-domain LRS3 and out-of-domain LRS2 benchmarks under clean and noise-corrupted conditions. All models were fine-tuned on the LRS3 433h dataset. For the noise fine-tuning setting, we applied randomly sampled MUSAN noise at a 0 dB SNR to prevent overfitting to specific noise types and ensure fair evaluation. Performance was assessed using WER and RERRs across three noise types---Babble, Music, and Natural---each tested at 10, 5, and 0 dB SNR levels.

The results clearly demonstrated the effectiveness of our routing mechanism: on the in-domain LRS3 test set, our method reduced the average WER from 13.43\% to 7.70\% under clean fine-tuning, yielding an RERR of 42.67\%. Under noisy conditions, WERs dropped significantly—for example, from 35.92\% to 21.40\% under 0 dB Babble noise. With noise-aware fine-tuning, the average WER decreased from 8.60\% to 7.18\%, corresponding to an RERR of 16.51\%, demonstrating the robustness of our approach across diverse acoustic conditions.

Notably, our approach also generalized well to the out-of-domain LRS2 set, which was unseen during training. Under clean fine-tuning, the average WER was reduced from 23.54\% to 19.49\%, yielding a RERR of 17.20\%. With noise fine-tuning, average WER decreased from 19.00\% to 18.40\% (RERR: 3.15\%). These consistent improvements across both in-domain and out-of-domain settings, and across various noise types and SNR levels, confirmed that the proposed router enhances AV-HuBERT's robustness against both additive noise and domain shift.

\subsection{Compatibility of AVFF Router with External Models}
To further demonstrate the modularity and broad applicability of our AVFF-based router, we evaluated its performance when integrated into Whisper-Flamingo, a representative multimodal AVSR system that differs architecturally from AV-HuBERT.
As shown in Table~\ref{tab:ood}, the AVFF router consistently improved performance across both in-domain and out-of-domain scenarios.
On the in-domain LRS3 test set, our method reduced the average WER from 4.49\% to 3.85\%, corresponding to an RERR of 14.25\%.
Similarly, on the out-of-domain LRS2 test set, average WER is reduced from 16.30\% to 15.24\%, achieving a 6.50\% RERR.
These results indicate that the proposed router functions as a general-purpose, plug-and-play module that can enhance noise robustness even when applied to external ASR backbones beyond AV-HuBERT.




\begin{figure} 
    \centering
    \includegraphics[width=\linewidth]{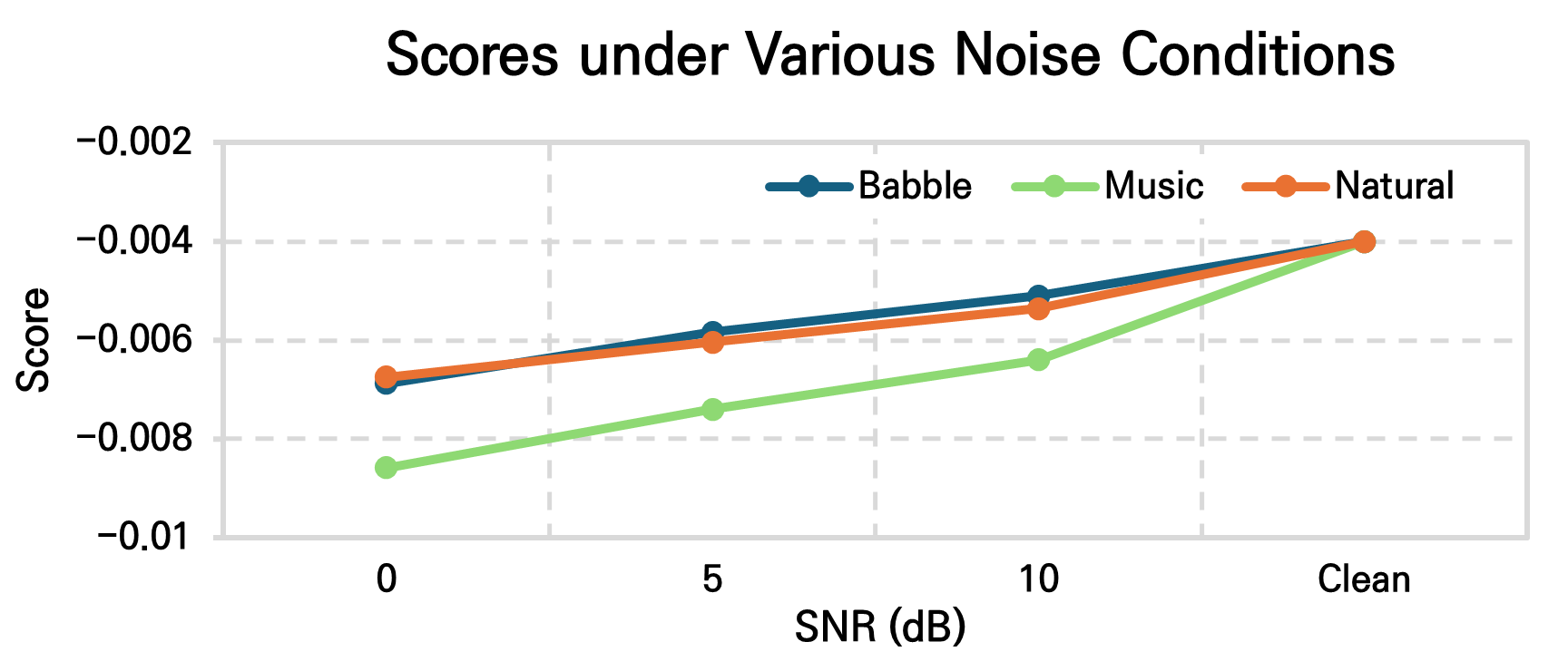}
    \caption{Average frame-level $\mathbf{s}_v$ scores under different noise types (Babble, Music, Natural) and SNR levels (0, 5, 10 dB, Clean), computed over all test samples per condition.}
    
    
    \label{figure_3}
    \vspace{-0.5em}
\end{figure}
\vspace{-0.5em}
\subsection{Analysis of Router Scores}
\vspace{-0.3em}
Fig.~\ref{figure_3} illustrates the average of frame-level scores ($\mathbf{s}_v$) generated by the proposed router on the LRS3 test set across various noise types (Babble, Music, Natural) and SNR levels (0, 5, 10, Clean dB). As the SNR decreases from clean to 0 dB, a consistent decline in $\mathbf{s}_v$ scores is observed across all noise types. This consistent downward trend indicates that the proposed router effectively captures the degradation of the audio modality and estimates token-level audio reliability under varying acoustic conditions.
\vspace{-0.3em}

\subsection{Ablation Results}
Table~\ref{tab:abls} presents the results of our ablation study, investigating the contribution of key components in our proposed architecture. We compared several model variants against our full model ('Ours') and a baseline. The evaluated variants are defined as follows:
\begin{itemize}
\item `$\mathbf{s}_a$': Replaces video score $\mathbf{s}_v$ with audio score $\mathbf{s}_a$.
\item `L2': Uses the L2 distance instead of the cosine similarity (Cos. Sim.) without complementarity.
\item `self-attn': Removes the GCA blocks and substitutes them with self-attention layers
\end{itemize} 
These experiments were conducted on the 30-hour subset of the LRS3 dataset using MUSAN noise at varying SNR levels. As presented in Table~\ref{tab:abls}, the AV-HuBERT baseline system added with only self-attention layers achieved an average WER of 7.24\%. Incorporating each component of our proposed method yielded consistent improvements, ultimately reducing the average WER to 6.67\% and demonstrating the complementary contribution of each module. Notably, the performance gains were observed not only on noisy datasets but also on clean data, highlighting the robustness and domain-generalization capability of our adaptive modality reweighting mechanism. 


\vspace{-1em}

\section{Conclusion}
In this paper, we presented a noise-aware decoding framework for fine-grained modality control, leveraging reliability scores from a pretrained router to balance audio and visual cues, outperforming AV-HuBERT on LRS3 and integrating with Whisper-Flamingo. We will investigate additional reliability signals in future work and evaluate against recent AVSR models to enhance generalization.

\section{Acknowledgement}
This work was partly supported by the National Research Foundation of Korea(NRF) grant funded by the Korea government(MSIT) (RS-2025-00557944) and Institute of Information \& communications Technology Planning \& Evaluation (IITP) grant funded by the Korea government(MSIT) (No.RS-2020-II201373, Artificial Intelligence Graduate School Program(Hanyang University))

\bibliographystyle{IEEEtran}
\bibliography{IEEEtranBST2/ref}

\begin{thebibliography}{10}
\providecommand{\url}[1]{#1}
\csname url@samestyle\endcsname
\providecommand{\newblock}{\relax}
\providecommand{\bibinfo}[2]{#2}
\providecommand{\BIBentrySTDinterwordspacing}{\spaceskip=0pt\relax}
\providecommand{\BIBentryALTinterwordstretchfactor}{4}
\providecommand{\BIBentryALTinterwordspacing}{\spaceskip=\fontdimen2\font plus
\BIBentryALTinterwordstretchfactor\fontdimen3\font minus
  \fontdimen4\font\relax}
\providecommand{\BIBforeignlanguage}[2]{{%
\expandafter\ifx\csname l@#1\endcsname\relax
\typeout{** WARNING: IEEEtran.bst: No hyphenation pattern has been}%
\typeout{** loaded for the language `#1'. Using the pattern for}%
\typeout{** the default language instead.}%
\else
\language=\csname l@#1\endcsname
\fi
#2}}
\providecommand{\BIBdecl}{\relax}
\BIBdecl

\bibitem{lin2025uncovering}
Z.~Lin and N.~Harte, ``Uncovering the visual contribution in audio-visual
  speech recognition,'' in \emph{Proc. IEEE International Conference on
  Acoustics, Speech and Signal Processing \textup{(ICASSP)}}, 2025.

\bibitem{avhubert}
B.~Shi, W.-N. Hsu, K.~Lakhotia, and A.~Mohamed, ``Learning audio-visual speech
  representation by masked multimodal cluster prediction,'' in \emph{Proc.
  International Conference on Learning Representations \textup{(ICLR)}}, 2022.

\bibitem{whisper-flamingo}
A.~Rouditchenko \emph{et~al.}, ``\textup{Whisper-Flamingo}: Integrating visual
  features into whisper for audio-visual speech recognition and translation,''
  in \emph{Proc. Interspeech}, 2024.

\bibitem{watchorlisten}
J.~Hong, M.~Kim, J.~Y. Choi, and Y.~M. Ro, ``Watch or listen: Robust
  audio-visual speech recognition with visual corruption modeling and
  reliability scoring,'' in \emph{Proc. IEEE/CVF Conference on Computer Vision
  and Pattern Recognition \textup{(CVPR)}}, 2023.

\bibitem{nevertolate}
C.~Chen \emph{et~al.}, ``It’s never too late: Fusing acoustic information
  into large language models for automatic speech recognition,'' in \emph{Proc.
  International Conference on Learning Representations \textup{(ICLR)}}, 2024.

\bibitem{snr_pred}
C.~Simic, K.~Riedhammer, and T.~Bocklet, ``Adapter-based multi-agent
  \textup{AVSR} extension for pre-trained \textup{ASR} models,'' in \emph{Proc.
  IEEE International Conference on Acoustics, Speech and Signal Processing
  \textup{(ICASSP)}}, 2025.

\bibitem{avff}
T.~Oorloff \emph{et~al.}, ``\textup{AVFF}: Audio-visual feature fusion for
  video deepfake detection,'' in \emph{Proc. IEEE/CVF Conference on Computer
  Vision and Pattern Recognition \textup{(CVPR)}}, 2024.

\bibitem{avsris}
D.~Serdyuk, O.~Braga, and O.~Siohan, ``Audio-visual speech recognition is worth
  32x32x8 voxels,'' in \emph{Proc. IEEE Automatic Speech Recognition and
  Understanding Workshop \textup{(ASRU)}}, 2021, pp. 796--802.

\bibitem{RNNT}
T.~Makino \emph{et~al.}, ``Recurrent neural network transducer for audio-visual
  speech recognition,'' in \emph{Proc. IEEE Automatic Speech Recognition and
  Understanding Workshop \textup{(ASRU)}}, 2019, pp. 905--912.

\bibitem{avdata2vec}
J.~Lian, A.~Baevski, W.-N. Hsu, and M.~Auli, ``\textup{AV-Data2Vec}:
  Self-supervised learning of audio-visual speech representations with
  contextualized target representations,'' in \emph{Proc. IEEE Automatic Speech
  Recognition and Understanding Workshop \textup{(ASRU)}}, 2023, pp. 1--8.

\bibitem{robustssl}
B.~Shi, W.-N. Hsu, and A.~Mohamed, ``Robust self-supervised audio-visual speech
  recognition,'' in \emph{Proc. Interspeech}, 2022, pp. 2118--2122.

\bibitem{noise}
D.~Snyder, G.~Chen, and D.~Povey, ``{MUSAN}: A music, speech, and noise
  corpus,'' \emph{ArXiv}, vol. abs/1510.08484, 2015.

\bibitem{ESC}
K.~J. Piczak, ``\textup{ESC}: Dataset for environmental sound classification,''
  in \emph{Proc. 23rd ACM International Conference on Multimedia
  \textup{(ACM)}}, 2015, pp. 1015--1018.

\bibitem{specaugment}
D.~S. Park \emph{et~al.}, ``Spec{A}ugment: A simple data augmentation method
  for automatic speech recognition,'' in \emph{Proc. Interspeech}, 2019.

\bibitem{dcimavsr}
X.~Wang \emph{et~al.}, ``{DCIM-AVSR}: Efficient audio-visual speech recognition
  via dual conformer interaction module,'' in \emph{Proc. IEEE International
  Conference on Acoustics, Speech and Signal Processing \textup{(ICASSP)}},
  2025.

\bibitem{DCCRN}
Y.~Hu \emph{et~al.}, ``\textup{DCCRN}: Deep complex convolution recurrent
  network for phase-aware speech enhancement,'' in \emph{Proc. Interspeech},
  2020.

\bibitem{demucs}
A.~Défossez, G.~Synnaeve, and Y.~Adi, ``Real time speech enhancement in the
  waveform domain,'' in \emph{Proc. Interspeech}, 2020.

\bibitem{avsrconformers}
P.~Ma, S.~Petridis, and M.~Pantic, ``End-to-end audio-visual speech recognition
  with conformers,'' in \emph{Proc. IEEE International Conference on Acoustics,
  Speech and Signal Processing \textup{(ICASSP)}}, 2021, pp. 7613--7617.

\bibitem{weighting}
H.~Glotin, D.~Vergyr, C.~Neti, G.~Potamianos, and J.~Luettin, ``Weighting
  schemes for audio-visual fusion in speech recognition,'' in \emph{Proc. IEEE
  International Conference on Acoustics, Speech and Signal Processing
  \textup{(ICASSP)}}, 2001.

\bibitem{unibev}
S.~Wang, H.~Caesar, L.~Nan, and J.~F. Kooij, ``Unibev: Multi-modal 3d object
  detection with uniform bev encoders for robustness against missing sensor
  modalities,'' in \emph{Proc. IEEE Intelligent Vehicles Symposium
  \textup{(IV)}}, 2024, pp. 2776--2783.

\bibitem{mcgurk}
H.~McGurk and J.~MacDonald, ``Hearing lips and seeing voices,'' \emph{Nature},
  vol. 264, no. 5588, pp. 746--748, 1976.

\bibitem{flexible}
E.~A.~B. Horrocks, F.~R. Rodrigues, and A.~B. Saleem, ``Flexible neural
  population dynamics govern the speed and stability of sensory encoding in
  mouse visual cortex,'' \emph{Nature Communications}, vol.~15, p. 6415, 2024.

\bibitem{align_linearinterpolation}
F.~Tao and C.~Busso, ``Aligning audiovisual features for audiovisual speech
  recognition,'' in \emph{Proc. IEEE International Conference on Multimedia and
  Expo \textup{(ICME)}}, 2018, pp. 1--6.

\bibitem{flamingo}
J.-B. Alayrac \emph{et~al.}, ``Flamingo: A visual language model for few-shot
  learning,'' in \emph{Proc. Advances in Neural Information Processing Systems
  \textup{(NeurIPS)}}, 2022.

\bibitem{layernorm}
J.~L. Ba, J.~R. Kiros, and G.~E. Hinton, ``Layer normalization,''
  \emph{arXiv:1607.06450}, 2016.

\bibitem{wgan}
M.~Arjovsky, S.~Chintala, and L.~Bottou, ``Wasserstein generative adversarial
  networks,'' in \emph{Proc. International conference on machine
  learning}.\hskip 1em plus 0.5em minus 0.4em\relax PMLR, 2017, pp. 214--223.

\bibitem{Adam}
D.~P. Kingma and J.~Ba, ``Adam: A method for stochastic optimization,''
  \emph{CoRR}, vol. abs/1412.6980, 2014.

\bibitem{lrs3}
T.~Afouras, J.~S. Chung, and A.~Zisserman, ``Lrs3-ted: a large-scale dataset
  for visual speech recognition,'' \emph{arXiv preprint arXiv:1809.00496},
  2018.

\bibitem{lrs2}
T.~Afouras, J.~S. Chung, A.~Senior, O.~Vinyals, and A.~Zisserman, ``Deep
  audio-visual speech recognition,'' \emph{IEEE Transactions on Pattern
  Analysis and Machine Intelligence \textup{(TPAMI)}}, 2018.

\bibitem{voxceleb}
A.~Nagrani, J.~S. Chung, and A.~Zisserman, ``Vox{C}eleb2: Deep speaker
  recognition.'' in \emph{Proc. Interspeech}, 2018.

\bibitem{MARLIN}
Z.~Cai \emph{et~al.}, ``\textup{MARLIN}: Masked autoencoder for facial video
  representation learning,'' in \emph{Proc. IEEE/CVF Conference on Computer
  Vision and Pattern Recognition \textup{(CVPR)}}, 2023.

\bibitem{AudioMAE}
P.-Y. Huang \emph{et~al.}, ``Masked autoencoders that listen,'' in \emph{Proc.
  Advances in Neural Information Processing Systems \textup{(NeurIPS)}}, 2022.

\bibitem{adamW}
I.~Loshchilov and F.~Hutter, ``Decoupled weight decay regularization,'' in
  \emph{Proc. International Conference on Learning Representations
  \textup{(ICLR)}}, 2018.

\end{thebibliography}

\end{document}